\documentclass[11pt,a4paper]{article}
\usepackage[hyperref]{acl2019}
\usepackage{times}
\usepackage{latexsym}
\usepackage{xcolor}
\usepackage{url}
\usepackage{graphicx}
\usepackage{multirow}
\usepackage{amsmath,,amsfonts,amssymb}
\usepackage{bm}
\usepackage{verbatim}
\usepackage{subcaption}
\usepackage[ruled,vlined,lined,linesnumbered]{algorithm2e}

\newcommand*\samethanks[1][\value{footnote}]{\footnotemark[#1]}

\newcommand{\approach}[1]{IRNet}
\newcommand{\lf}[1]{SemQL}

\aclfinalcopy 

\title{Towards Complex Text-to-SQL in Cross-Domain Database with Intermediate Representation}

\author{
Jiaqi Guo\textsuperscript{1}\thanks{\indent Equal Contributions. Work done during an internship at MSRA.},
Zecheng Zhan\textsuperscript{2}\samethanks,
Yan Gao\textsuperscript{3},
Yan Xiao\textsuperscript{3},
Jian-Guang Lou\textsuperscript{3}
\\{\bf Ting Liu\textsuperscript{1}, Dongmei Zhang\textsuperscript{3}}\\
  \textsuperscript{1}Xi'an Jiaotong University, Xi'an, China\\
  \textsuperscript{2}Beijing University of Posts and Telecommunications, Beijing, China\\
  \textsuperscript{3}Microsoft Research Asia, Beijing, China\\
  {\tt jasperguo2013@stu.xjtu.edu.cn,zhanzecheng@bupt.edu.cn}\\
  {\tt \{Yan.Gao,Yan.Xiao,jlou,dongmeiz\}@microsoft.com}\\
  {\tt tingliu@mail.xjtu.edu.cn}\\
}

\date{}

\begin{document}
\maketitle
\begin{abstract}
We present a neural approach called \approach{} for complex and cross-domain Text-to-SQL.
\approach{} aims to address two challenges:
1) the mismatch between intents expressed in natural language (NL) and the implementation details in SQL;
2) the challenge in predicting columns caused by the large number of out-of-domain words.
Instead of end-to-end synthesizing a SQL query, \approach{} decomposes the synthesis process into three phases.
In the first phase, \approach{} performs a schema linking over a question and a database schema.
Then, \approach{} adopts a grammar-based neural model to synthesize
a \lf{} query which is an intermediate representation that we design to bridge NL and SQL.
Finally, \approach{} deterministically infers a SQL query from the synthesized \lf{} query with domain knowledge.
On the challenging Text-to-SQL benchmark Spider, \approach{} achieves $46.7\%$ accuracy, obtaining $19.5\%$ absolute improvement over previous state-of-the-art approaches.
At the time of writing, \approach{} achieves the first position on the Spider leaderboard.
\end{abstract}

\section{Introduction}
\label{sec:intro}

Recent years have seen a great deal of renewed interest in Text-to-SQL, i.e., synthesizing a SQL query from a question.
Advanced neural approaches synthesize SQL queries in an end-to-end manner and achieve more than $80\%$ exact matching accuracy on public Text-to-SQL benchmarks (e.g., ATIS, GeoQuery and WikiSQL)~\citep{Krishnamurthy2017NSP, ZhongSeq2SQL2017,Sqlnet2017, Sqlizer2018, Typesql2018, Dong2018coarse2fine, wang2018robust, Noval2019}.
However, \citet{Spider2018} presents unsatisfactory performance of state-of-the-art approaches on a newly released, cross-domain Text-to-SQL benchmark, Spider.

\begin{figure}[t!]
	\centering
	\includegraphics[scale=0.50]{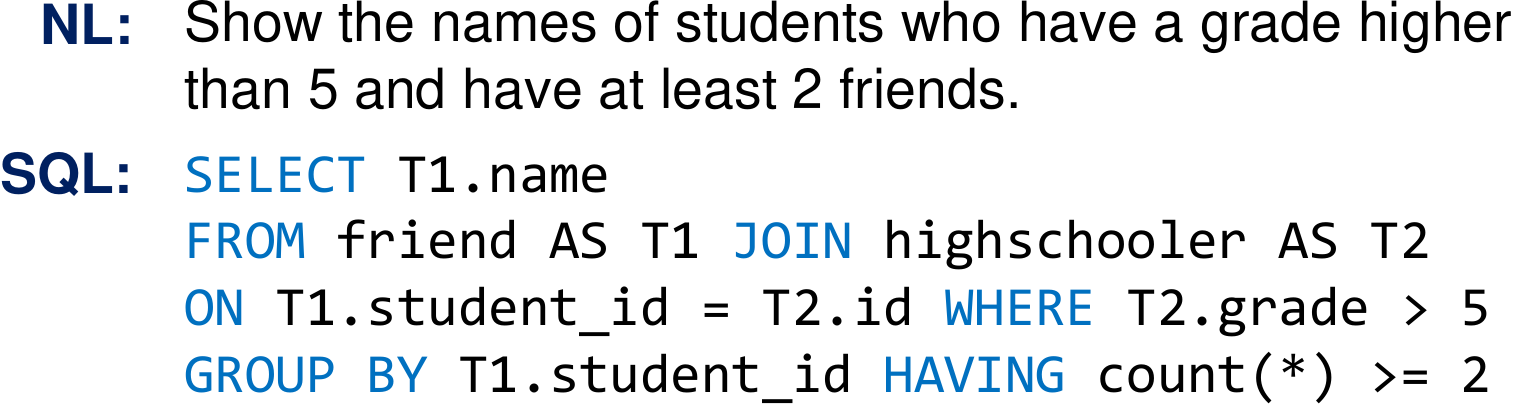}
	\caption{
		\label{fig:intro_example} An example from the Spider benchmark to illustrate the mismatch between the intent expressed in NL and the implementation details in SQL. The column \textsl{`student\_id'} to be grouped by in the SQL query is not mentioned in the question.}
\end{figure}

The Spider benchmark brings new challenges that prove to be hard for existing approaches.
Firstly, the SQL queries in the Spider contain nested queries and clauses like \texttt{GROUPBY} and \texttt{HAVING}, which are far more complicated than that in another well-studied cross-domain benchmark, WikiSQL~\citep{ZhongSeq2SQL2017}.
Considering the example in Figure~\ref{fig:intro_example}, the column \textsl{`student\_id'} to be grouped by in the SQL query is never mentioned in the question.
In fact, the \texttt{GROUPBY} clause is introduced in SQL to facilitate the implementation of aggregate functions.
Such implementation details, however, are rarely considered by end users and therefore rarely mentioned in questions.
This poses a severe challenge for existing end-to-end neural approaches to synthesize SQL queries in the absence of detailed specification.
The challenge in essence stems from the fact that SQL is designed for effectively querying relational databases instead of for representing the meaning of NL~\cite{kate2008transforming}.
Hence, there inevitably exists a mismatch between intents expressed in natural language and the implementation details in SQL. 
We regard this challenge as a mismatch problem.

Secondly, given the cross-domain settings of Spider, there are a large number of out-of-domain (OOD) words.
For example, $35\%$ of words in database schemas on the development set do not occur in the schemas on the training set in Spider.
As a comparison, the number in WikiSQL is only $22\%$.
The large number of OOD words poses another steep challenge in predicting columns in SQL queries~\citep{Syntaxsqlnet2018}, because the OOD words usually lack of accurate representations in neural models.
We regard this challenge as a lexical problem.

In this work, we propose a neural approach, called \approach{}, towards tackling the mismatch problem and the lexical problem with intermediate representation and schema linking.
Specifically, instead of end-to-end synthesizing a SQL query from a question, \approach{} decomposes the synthesis process into three phases.
In the first phase, \approach{} performs a schema linking over a question and a schema.
The goal of the schema linking is to recognize the columns and the tables mentioned in a question, and to assign different types to the columns based on how they are mentioned in the question.
Incorporating the schema linking can enhance the representations of question and schema, especially when the OOD words lack of accurate representations in neural models during testing.
Then, \approach{} adopts a grammar-based neural model to synthesize a \lf{} query, which is an intermediate representation (IR) that we design to bridge NL and SQL.
Finally, \approach{} deterministically infers a SQL query from the synthesized \lf{} query with domain knowledge.

The insight behind \approach{} is primarily inspired by the success of using intermediate representations (e.g., lambda calculus~\citep{carpenter1997type}, FunQL~\cite{Kate2005LTN} and DCS~\citep{DCS2011}) in various semantic parsing tasks~\cite{Zelle1996LPD, Berant2013Freebase, Compositional2015Pasupat, Wang2017SHE},
and previous attempts in designing IR to decouple meaning representations of NL from database schema and database management system~\citep{Woods1986Lunar, alshawi1992core, Androutsopoulos1993MASQUE}.

On the challenging Spider benchmark~\citep{Spider2018}, \approach{} achieves $46.7\%$ exact matching accuracy, obtaining $19.5\%$ absolute improvement over previous state-of-the-art approaches.
At the time of writing, \approach{} achieves the first position on the Spider leaderboard.
When augmented with BERT~\citep{Bert2018}, \approach{} reaches up to $54.7\%$ accuracy.
In addition, as we show in the experiments, learning to synthesize \lf{} queries rather than SQL queries can substantially benefit other neural approaches for Text-to-SQL, such as SQLNet~\citep{Sqlnet2017}, TypeSQL~\citep{Typesql2018} and SyntaxSQLNet~\citep{Syntaxsqlnet2018}.
Such results on the one hand demonstrate the effectiveness of \lf{} in bridging NL and SQL.
On the other hand, it reveals that designing an effective intermediate representation to bridge NL and SQL is a promising direction to being there for complex and cross-domain Text-to-SQL.

\section{Approach}
\label{sec:methodology}

In this section, we present \approach{} in detail.
We first describe how to tackle the mismatch problem and the lexical problem with intermediate representation and schema linking.
Then we present the neural model to synthesize \lf{} queries.

\subsection{Intermediate Representation}
\label{sec:lf}

\begin{figure}[t!]
	\centering
	\includegraphics[scale=0.42]{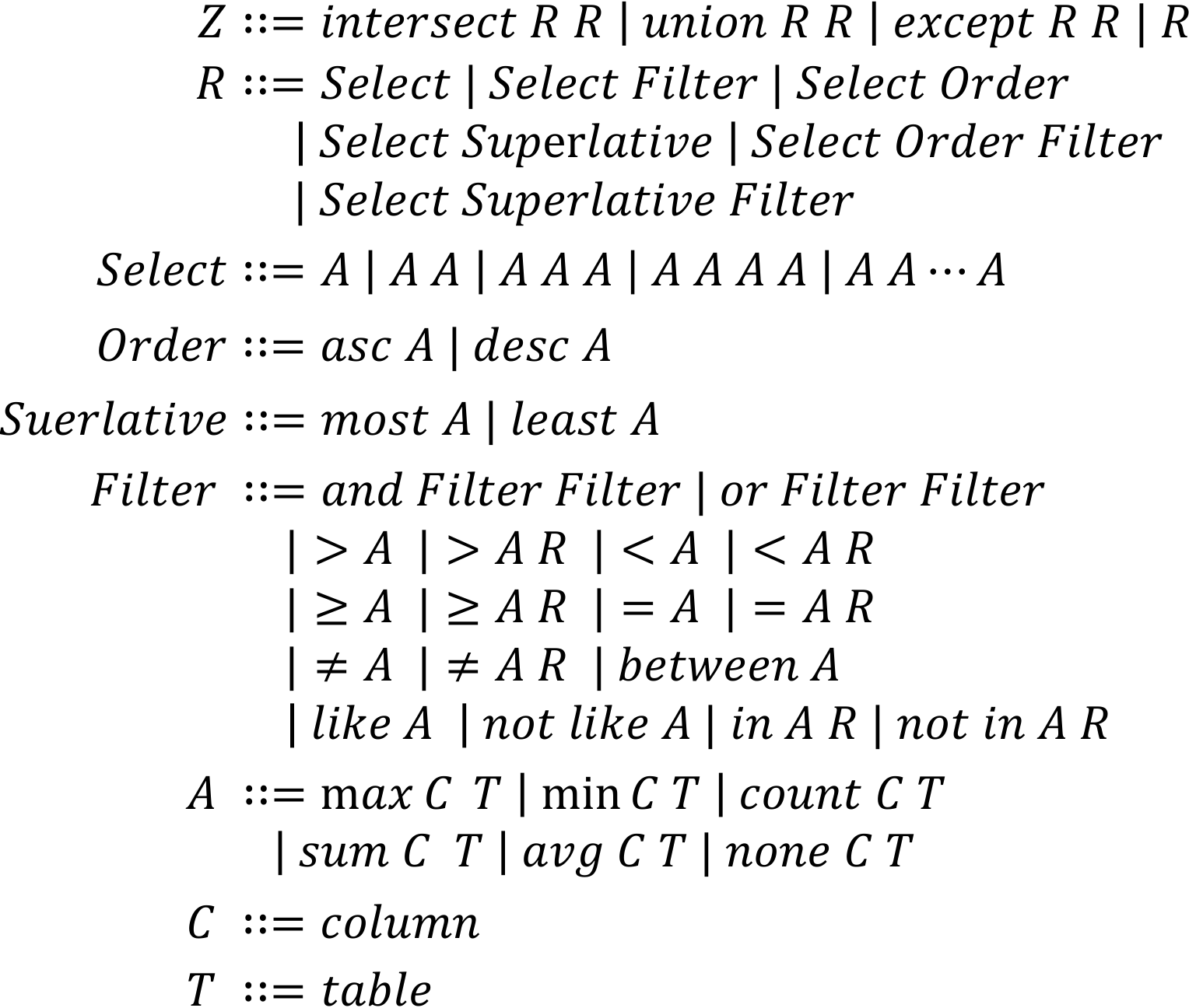}
	\caption{
		\label{fig:grammar} The context-free grammar of \lf{}. $column$ ranges over distinct column names in a schema. $table$ ranges over tables in a schema.}
\end{figure}

\begin{figure}[t!]
	\centering
	\includegraphics[scale=0.48]{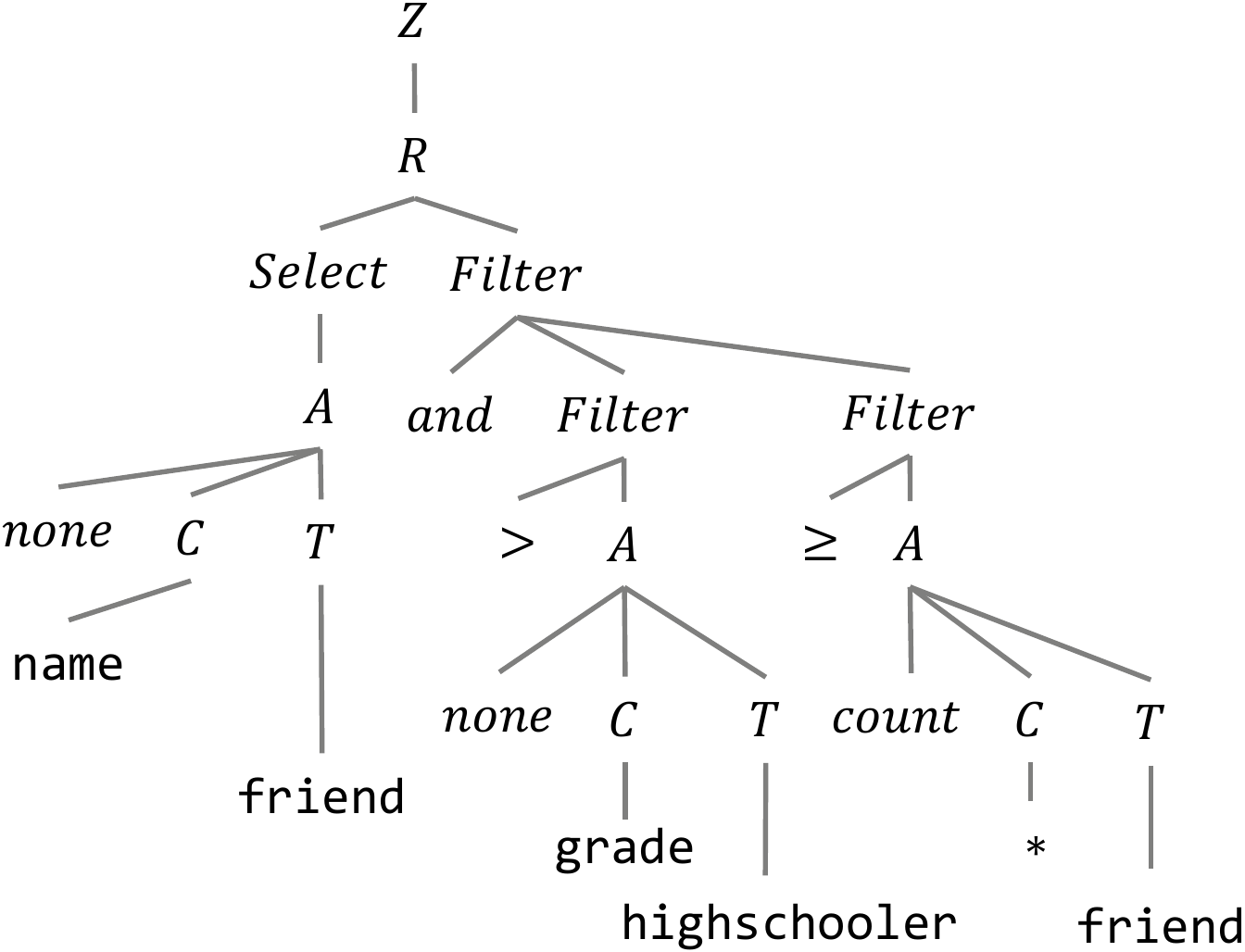}
	\caption{
		\label{fig:lf_example} An illustrative example of \lf{}. Its corresponding question and SQL query are shown in Figure~\ref{fig:intro_example}.}
\end{figure}

To eliminate the mismatch, we design a domain specific language, called \lf{}, which serves as an intermediate representation between NL and SQL.
Figure~\ref{fig:grammar} presents the context-free grammar of \lf{}.
An illustrative \lf{} query is shown in Figure~\ref{fig:lf_example}. 
We elaborate on the design of \lf{} in the following.

Inspired by lambda DCS~\citep{liang2013lambda}, \lf{} is designed to be tree-structured.
This structure, on the one hand, can effectively constrain the search space during synthesis.
On the other hand, in view of the tree-structure nature of SQL~\cite{Syntaxsqlnet2018,Yin2018TRANX}, following the same structure also makes it easier to translate to SQL intuitively.

The mismatch problem is mainly caused by the implementation details in SQL queries and missing specification in questions as discussed in Section~\ref{sec:intro}.
Therefore, it is natural to hide the implementation details in the intermediate representation, which forms the basic idea of \lf{}.
Considering the example in Figure~\ref{fig:lf_example}, the \texttt{GROUPBY}, \texttt{HAVING} and \texttt{FROM} clauses in the SQL query are eliminated in the \lf{} query, and the conditions in \texttt{WHERE} and \texttt{HAVING} are uniformly expressed in the subtree of \textit{Filter} in the \lf{} query.
The implementation details can be deterministically inferred from the \lf{} query in the later inference phase with domain knowledge.
For example, a column in the \texttt{GROUPBY} clause of a SQL query usually occurs in the \texttt{SELECT} clause or it is the primary key of a table where an aggregate function is applied to one of its columns.

In addition, we strictly require to declare the table that a column belongs to in \lf{}.
As illustrated in Figure~\ref{fig:lf_example}, the column \textsl{`name'} along with its table \textsl{`friend'} are declared in the \lf{} query.
The declaration of tables helps to differentiate duplicated column names in the schema.
We also declare a table for the special column `\texttt{*}' because we observe that `\texttt{*}' usually aligns with a table mentioned in a question.
Considering the example in Figure~\ref{fig:lf_example}, the column `\texttt{*}' in essence aligns with the table \textsl{`friend'}, which is explicitly mentioned in the question.
Declaring a table for `\texttt{*}' also helps infer the \texttt{FROM} clause in the next inference phase.

When it comes to inferring a SQL query from a \lf{} query, we perform the inference based on an assumption that the definition of a database schema is precise and complete.
Specifically, if a column is a foreign key of another table, there should be a foreign key constraint declared in the schema.
This assumption usually holds as it is the best practice in database design.
More than $95\%$ of examples in the training set of the Spider benchmark hold this assumption.
The assumption forms the basis of the inference.
Take the inference of the \texttt{FROM} clause in a SQL query as an example.
We first identify the shortest path that connects all the declared tables in a \lf{} query in the schema (A database schema can be formulated as an undirected graph, where vertex are tables and edges are foreign key relations among tables).
Joining all the tables in the path eventually builds the \texttt{FROM} clause.
Supplementary materials provide detailed procedures of the inference and more examples of \lf{} queries.

\subsection{Schema Linking}
The goal of schema linking in \approach{} is to recognize the columns and the tables mentioned in a question, and assign different types to the columns based on how they are mentioned in the question.
Schema linking is an instantiation of entity linking in the context of Text-to-SQL, where entity is referred to columns, tables and cell values in a database.
We use a simple yet effective string-match based method to implement the linking.
In the followings, we illustrate how \approach{} performs schema linking in details based on the assumption that the cell values in a database are not available.

\begin{figure*}[t!]
	\centering
	\includegraphics[scale=0.33]{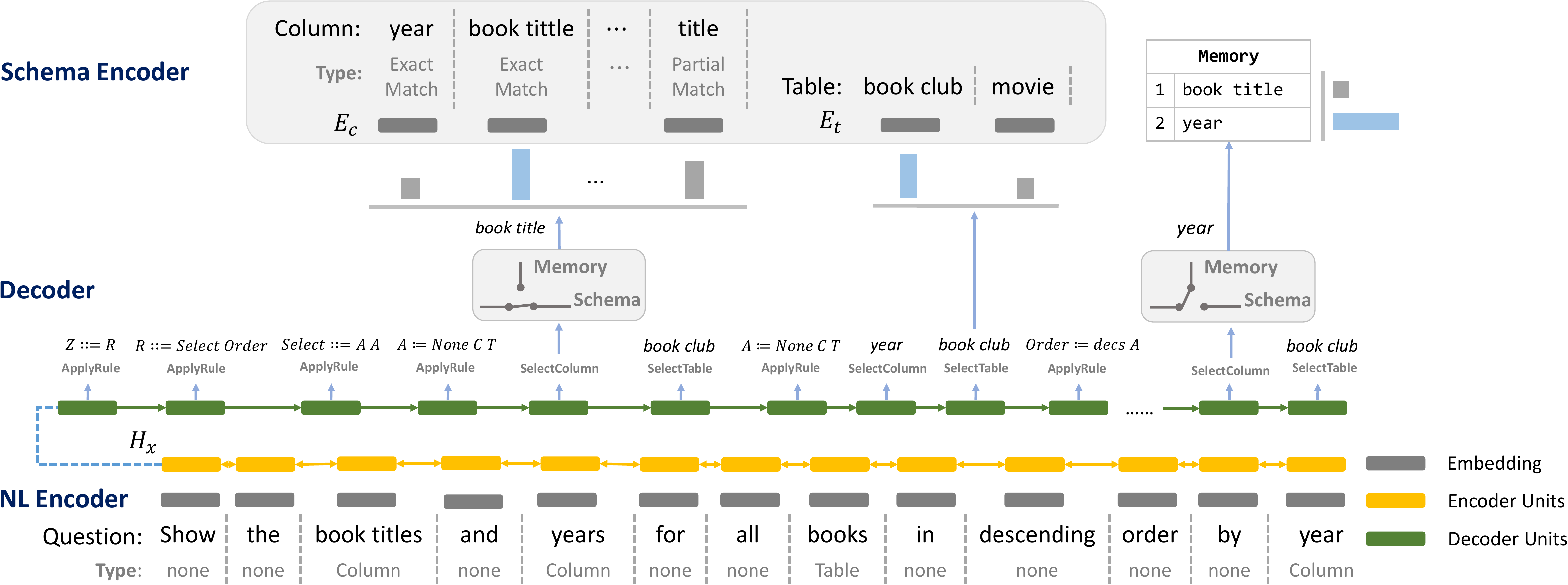}
	\caption{
		\label{fig:architecture} An overview of the neural model to synthesize \lf{} queries. Basically, \approach{} is constituted by an NL encoder, a schema encoder and a decoder. As shown in the figure, the column `book title' is selected from the schema, while the second column `year' is selected from the memory.}
\end{figure*}

As a whole, we define three types of entities that may be mentioned in a question, namely, \textit{table}, \textit{column} and \textit{value}, where \textit{value} stands for a cell value in the database.
In order to recognize entities, we first enumerate all the n-grams of length 1-6 in a question.
Then, we enumerate them in the descending order of length.
If an n-gram exactly matches a column name or is a subset of a column name, we recognize this n-gram as a \textit{column}. 
The recognition of \textit{table} follows the same way.
If an n-gram can be recognized as both \textit{column} and \textit{table}, we prioritize \textit{column}.
If an n-gram begins and ends with a single quote, we recognize it as \textit{value}.
Once an n-gram is recognized, we will remove other n-grams that overlap with it.
To this end, we can recognize all the entities mentioned in a question and obtain a non-overlap n-gram sequence of the question by joining those recognized n-grams and the remaining 1-grams.
We refer each n-gram in the sequence as a span and assign each span a type according to its entity.
For example, if a span is recognized as \textit{column}, we will assign it a type \textsc{Column}.
Figure~\ref{fig:architecture} depicts the schema linking results of a question.

For those spans recognized as \textit{column}, if they exactly match the column names in the schema, we assign these columns a type \textsc{Exact Match}, otherwise a type \textsc{Partial Match}. 
To link the cell value with its corresponding column in the schema, we first query the \textit{value} span in ConceptNet~\citep{ConceptNet2012} which is an open, large-scale knowledge graph and search the results returned by ConceptNet over the schema.
We only consider the query results in two categories of ConceptNet, namely, `is a type of' and `related terms', as we observe that the column that a cell value belongs to usually occurs in these two categories.
If there exists a result exactly or partially matches a column name in the schema, we assign the column a type \textsc{Value Exact Match} or \textsc{Value Partial Match}.
\subsection{Model}
\label{sec:approach}

We present the neural model to synthesize \lf{} queries, which takes a question, a database schema and the schema linking results as input.
Figure~\ref{fig:architecture} depicts the overall architecture of the model via an illustrative example.

To address the lexical problem, we consider the schema linking results when constructing representations for the question and columns in the schema.
In addition, we design a memory augmented pointer network for selecting columns during synthesis.
When selecting a column, it makes a decision first on whether to select from memory or not, which sets it apart from the vanilla pointer network~\citep{Vinvals2015Pointer}.
The motivation behind the memory augmented pointer network is that the vanilla pointer network is prone to selecting same columns according to our observations.

\noindent \textbf{NL Encoder.}
Let $x $$=$$ [(x_1, \tau_1), \cdots, (x_L, \tau_L)]$ denote the non-overlap span sequence of a question, where $x_i$ is the $i^{th}$ span and $\tau_i$ is the type of span $x_i$ assigned in schema linking.
The NL encoder takes $x$ as input and encodes $x$ into a sequence of hidden states $\bm{H}_x$.
Each word in $x_i$ is converted into its embedding vector and its type $\tau_i$ is also converted into an embedding vector.
Then, the NL encoder takes the average of the type and word embeddings as the span embedding $\bm{e}^i_{x}$.
Finally, the NL encoder runs a bi-directional LSTM~\citep{hochreiter1997long} over all the span embeddings.
The output hidden states of the forward and backward LSTM are concatenated to construct $\bm{H}_x$.

\noindent \textbf{Schema Encoder.}
Let $s $$=$$ (c, t)$ denote a database schema, where $c $$=$$ \{(c_1, \phi_i), \cdots, (c_n, \phi_n)\}$ is the set of distinct columns and their types that we assign in schema linking, and $t $$=$$ \{t_1, \cdots, t_m\}$ is the set of tables.
The schema encoder takes $s$ as input and outputs representations for columns $\bm{E}_c$ and tables $\bm{E}_t$.
We take the column representations as an example below.
The construction of table representations follows the same way except that we do not assign a type to a table in schema linking.

Concretely, each word in $c_i$ is first converted into its embedding vector and its type $\phi_i$ is also converted into an embedding vector $\bm{\varphi}_i$.
Then, the schema encoder takes the average of word embeddings as the initial representations $\bm{\hat{e}}^i_c$ for the column.
The schema encoder further performs an attention over the span embeddings and obtains a context vector $\bm{c}^i_c$.
Finally, the schema encoder takes the sum of the initial embedding, context vector and the type embedding as the column representation $\bm{e}^i_c$.
The calculation of the representations for column $c_i$ is as follows.
\begin{equation*}
\begin{aligned}
g^i_k &= \dfrac{(\bm{\hat{e}}^i_c)^\mathsf{T} \bm{e}^k_x}{\Vert \bm{\hat{e}}^i_c \Vert \Vert \bm{e}^k_x \Vert} \\
\bm{c}^i_c &= \sum^{L}_{k=1}g^i_k\bm{e}^k_x \\
\bm{e}^i_c &= \bm{\hat{e}}^i_c + \bm{c}^i_c + \bm{\varphi}_i,
\end{aligned}
\end{equation*}


\noindent \textbf{Decoder.} 
The goal of the decoder is to synthesize \lf{} queries.
Given the tree structure of \lf{}, we use a grammar-based decoder~\cite{Yin2017Grammar, Yin2018TRANX} which leverages a LSTM to model the generation process of a \lf{} query via sequential applications of actions.
Formally, the generation process of a \lf{} query $y$ can be formalized as follows.
\begin{equation*}
	\begin{aligned}
		p(y|x, s) &= \prod^T_{i=1}p(a_i|x, s, a_{<i}),
	\end{aligned}
\end{equation*}
where $a_i$ is an action taken at time step $i$, $a_{<i}$ is the sequence of actions before $i$, and $T$ is the number of total time steps of the whole action sequence.

The decoder interacts with three types of actions to generate a \lf{} query, including \textsc{ApplyRule}, \textsc{SelectColumn} and \textsc{SelectTable}.
\textsc{ApplyRule}($r$) applies a production rule $r$ to the current derivation tree of a \lf{} query.
\textsc{SelectColumn}($c$) and \textsc{SelectTable}($t$) selects a column $c$ and a table $t$ from the schema, respectively.
Here, we detail the action \textsc{SelectColumn} and \textsc{SelectTable}. 
Interested readers can refer to ~\citet{Yin2017Grammar} for details of the action \textsc{ApplyRule}. 

We design a memory augmented pointer network to implement the action \textsc{SelectColumn}.
The memory is used to record the selected columns, which is similar to the memory mechanism used in~\citet{Liang2017NSM}.
When the decoder is going to select a column, it first makes a decision on whether to select from the memory or not, and then selects a column from the memory or the schema based on the decision.
Once a column is selected, it will be removed from the schema and be recorded in the memory.
The probability of selecting a column $c$ is calculated as follows.
\begin{equation*}
\begin{aligned}
p(a_i&=\textsc{SelectColumn}[c]|x, s, a_{<i}) = \\
& p(\textsc{Mem}|x, s, a_{<i})p(c|x, s, a_{<i}, \textsc{Mem}) \\
& + p(\textsc{S}|x, s, a_{<i})p(c|x, s, a_{<i}, \textsc{S}) \\
\end{aligned}
\end{equation*}
\begin{equation*}
\begin{aligned}
p(\textsc{Mem}|x, s, a_{<i}) &= \mathrm{sigmod}(\bm{w}_m^\mathsf{T} \bm{v}_i) \\
p(\textsc{S}|x, s, a_{<i}) &= 1 - p(\textsc{Mem}|x, s, a_{<i})\\
p(c|x, s, a_{<i}, \textsc{Mem}) &\propto \exp(\bm{v}_i^\mathsf{T}\bm{E}^m_c)
\\
p(c|x, s, a_{<i}, \textsc{S}) &\propto \exp(\bm{v}_i^\mathsf{T}\bm{E}^s_c),
\end{aligned}
\end{equation*}
where \textsc{S} represents selecting from schema, \textsc{Mem} represents selecting from memory, $\bm{v}_i$ denotes the context vector that is obtained by performing an attention over $\bm{H}_x$, $\bm{E}^m_c$ denotes the embedding of columns in memory and $\bm{E}^s_c$ denotes the embedding of columns that are never selected.
$\bm{w}_m$ is trainable parameter.

When it comes to \textsc{SelectTable}, the decoder selects a table $t$ from the schema via a pointer network:
\begin{equation*}
\begin{aligned}
p(a_i=\textsc{SelectTable}[t]|x, s, a_{<i}) &\propto \exp(\bm{v}_i^\mathsf{T}\bm{E}_t).
\end{aligned}
\end{equation*}
As shown in Figure~\ref{fig:architecture}, the decoder first predicts a column and then predicts the table that it belongs to.
To this end, we can leverage the relations between columns and tables to prune the irrelevant tables.

\noindent \textbf{Coarse-to-fine.} We further adopt a coarse-to-fine framework~\citep{Armando2008Sketch, Bornholt2016Metasketches, Dong2018coarse2fine}, decomposing the decoding process of a \lf{} query into two stages.
In the first stage, a skeleton decoder outputs a skeleton of the \lf{} query.
Then, a detail decoder fills in the missing details in the skeleton by selecting columns and tables.
Supplementary materials provide a detailed description of the skeleton of a \lf{} query and the coarse-to-fine framework.

\section{Experiment}
\label{sec:experiment}

In this section, we evaluate the effectiveness of \approach{} by comparing it to the state-of-the-art approaches and ablating several design choices in \approach{} to understand their contributions.

\subsection{Experiment Setup}

\noindent \textbf{Dataset.}
We conduct our experiments on the Spider \citep{Spider2018}, a large-scale, human-annotated and cross-domain Text-to-SQL benchmark.
Following \citet{Syntaxsqlnet2018}, we use the database split for evaluations, where $206$ databases are split into $146$ training, $20$ development and $40$ testing.
There are $8625$, $1034$, $2147$ question-SQL query pairs for training, development and testing.
Just like any competition benchmark, the test set of Spider is not publicly available, and our models are submitted to the data owner for testing.
We evaluate \approach{} and other approaches using SQL Exact Matching and Component Matching proposed by \citet{Spider2018}.

\noindent \textbf{Baselines.}
We also evaluate the sequence-to-sequence model \citep{Seq2seq2014} augmented with a neural attention mechanism \cite{Attention2015} and a copying mechanism \citep{Copying2016}, SQLNet \cite{Sqlnet2017},  TypeSQL \cite{Typesql2018}, and SyntaxSQLNet \cite{Syntaxsqlnet2018} which is the state-of-the-art approach on the Spider.

\noindent \textbf{Implementations.}
We implement \approach{} and the baseline approaches with PyTorch \citep{Pytorch2017}.
Dimensions of word embeddings, type embeddings and hidden vectors are set to 300.
Word embeddings are initialized with Glove \citep{Glove2014} and shared between the NL encoder and schema encoder.
They are fixed during training.
The dimension of action embedding and node type embedding are set to 128 and 64, respectively. 
The dropout rate is 0.3.
We use Adam \citep{Adam2015} with default hyperparameters for optimization.
Batch size is set to 64.

\noindent \textbf{BERT.}
Language model pre-training has shown to be effective for learning universal language representations.
To further study the effectiveness of our approach, inspired by SQLova \cite{Noval2019}, we leverage BERT~\citep{Bert2018} to encode questions, database schemas and the schema linking results. 
The decoder remains the same as in \approach{}.
Specifically, the sequence of spans in the question are concatenated with all the distinct column names in the schema.
Each column name is separated with a special token \textit{[SEP]}.
BERT takes the concatenation as input.
The representation of a span in the question is taken as the average hidden states of its words and type.
To construct the representation of a column, we first run a bi-directional LSTM (BI-LSTM) over the hidden states of its words.
Then, we take the sum of its type embedding and the final hidden state of the BI-LSTM as the column representation.
The construction of table representations follows the same way.
Supplementary material provides a figure to illustrate the architecture of the encoder.
To establish baseline, we also augment SyntaxSQLNet with BERT.
Note that we only use the base version of BERT due to the resource limitations.

We do not perform any data augmentation for fair comparison.
All our code are publicly available.~\footnote{\href{https://github.com/zhanzecheng/IRNet}{https://github.com/zhanzecheng/IRNet}}


\begin{table}[t!]
	\small
	\centering
	\begin{tabular}{lcc}
		\hline 
		\multicolumn{1}{l}{\textbf{Approach}} & 
		\multicolumn{1}{c}{\textbf{Dev}}   & 
		\multicolumn{1}{c}{\textbf{Test}} \\ \hline \hline
		Seq2Seq & 1.9\% & 3.7\% \\
		Seq2Seq + Attention & 1.8\% & 4.8\% \\
		Seq2Seq + Copying & 4.1\% & 5.3\% \\
		TypeSQL & 8.0\% & 8.2\% \\ 
		SQLNet & 10.9\% & 12.4\% \\
		SyntaxSQLNet & 18.9\% & 19.7\% \\ 
		SyntaxSQLNet(augment) & 24.8\% & 27.2\% \\ 
		\textbf{\approach{}} & \textbf{53.2\%} & \textbf{46.7\%} \\
		\hline
		\textbf{BERT} & & \\
		SyntaxSQLNet(BERT) & 25.0\% & 25.4\% \\ 
		\textbf{\approach{}(BERT)} & \textbf{61.9\%} & \textbf{54.7\%} \\ \hline
	\end{tabular}
	\caption{\label{tab:main_results} Exact matching accuracy on SQL queries.}
\end{table}
\begin{figure}[t!]
	\centering
	\includegraphics[scale=0.35]{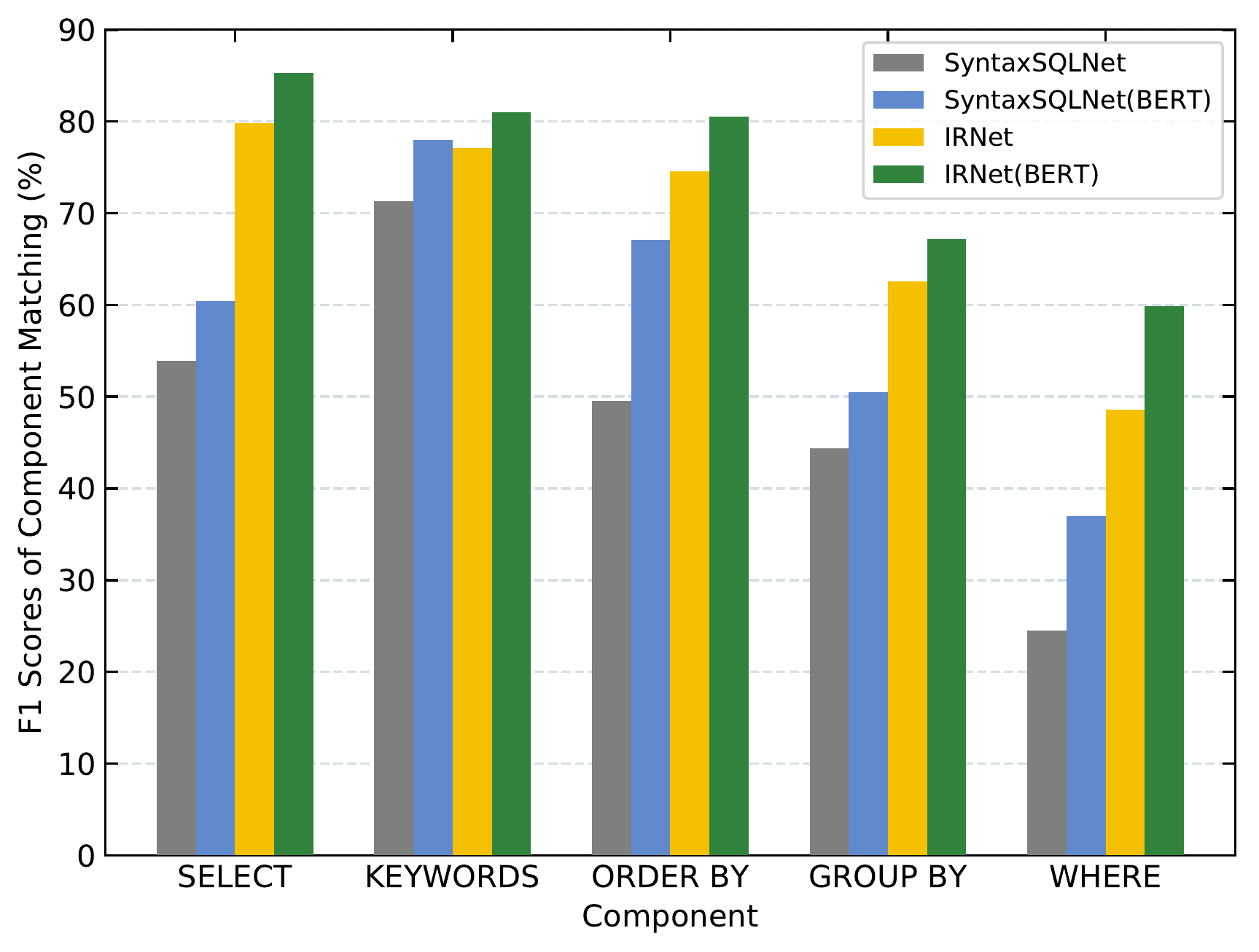}
	\caption{
		\label{fig:f1_by_component} F1 scores of component matching of SyntaxSQLNet, SyntaxSQLNet(BERT), \approach{} and \approach{}(BERT) on test set.}
\end{figure}

\subsection{Experimental Results}
Table~\ref{tab:main_results} presents the exact matching accuracy of \approach{} and various baselines on the development set and the test set.
\approach{} clearly outperforms all the baselines by a substantial margin.
It obtains $27.0\%$ absolute improvement over SyntaxSQLNet on test set.
It also obtains $19.5\%$ absolute improvement over SyntaxSQLNet(augment) that performs large-scale data augmentation.
When incorporating BERT, the performance of both SyntaxSQLNet and \approach{} is substantially improved and the accuracy gap between them on both the development set and the test set is widened.

To study the performance of \approach{} in detail, following~\citet{Syntaxsqlnet2018}, we measure the average F1 score on different SQL components on the test set.
We compare between SyntaxSQLNet and \approach{}.
As shown in Figure~\ref{fig:f1_by_component}, \approach{} outperforms SyntaxSQLNet on all components.
There are at least $18.2\%$ absolute improvement on each component except KEYWORDS.
When incorporating BERT, the performance of \approach{} on each component is further boosted, especially in \texttt{WHERE} clause.

We further study the performance of \approach{} on different portions of the test set according to the hardness levels of SQL defined in~\citet{Spider2018}.
As shown in Table~\ref{tab:accuracy_by_hardness}, \approach{} significantly outperforms SyntaxSQLNet in all four hardness levels with or without BERT.
For example, compared with SyntaxSQLNet, \approach{} obtains $23.3\%$ absolute improvement in Hard level.
\begin{table}[h]
	\small
	\centering
	\begin{tabular}{lrrrr}
		\hline 
		\small{\textbf{\multirow{2}*{Approach}}} & 
		\small{\textbf{\multirow{2}*{Easy}}}   & 
		\small{\textbf{\multirow{2}*{Medium}}}  & 
		\small{\textbf{\multirow{2}*{Hard}}}  & 
		\small{\textbf{Extra}} \\
		&&&&\small{\textbf{Hard}}\\\hline  \hline
		SyntaxSQLNet  & 38.6\% & 17.6\% & 16.3\% & 4.9\%  \\
        SyntaxSQLNet  & \multirow{2}*{42.9\%} & \multirow{2}*{24.9\%} & \multirow{2}*{21.9\%}& \multirow{2}*{8.6\%} \\ 
		\small{(BERT)}\\ \hline
		\textbf{\approach{}}  & \textbf{70.1\%} & \textbf{49.2\%} & \textbf{39.5\%} & \textbf{19.1\%} \\
		\textbf{\approach{}(BERT)}  & \textbf{77.2\%} & \textbf{58.7\%} & \textbf{48.1\%} & \textbf{25.3\%}  \\
		\hline
	\end{tabular}
	\caption{\label{tab:accuracy_by_hardness} Exact matching accuracy of SyntaxSQLNet, SyntaxSQLNet(BERT), \approach{} and \approach{}(BERT) on test set by hardness level.}
\end{table}

\begin{table}[t!]
	\small
	\centering
	\begin{tabular}{lcc}
		\hline 
		\multicolumn{1}{l}{\textbf{Approach}} & 
		\multicolumn{1}{c}{\textbf{SQL}} & 
		\multicolumn{1}{c}{\textbf{\lf{}}}  \\ \hline \hline
		Seq2Seq & 1.9\% & 11.4\%(\textbf{+9.5}) \\
		Seq2Seq + Attention & 1.8\% & 14.7\%(\textbf{+12.9}) \\
		Seq2Seq + Copying & 4.1\% & 18.5\%(\textbf{+14.1}) \\
		TypeSQL & 8.0\% & 14.4\%(\textbf{+6.4}) \\ 
		SQLNet & 10.9\% & 17.5\%(\textbf{+6.6}) \\
		SyntaxSQLNet & 18.9\% & 27.5\%(\textbf{+8.6}) \\ 
		\hline
		\textbf{BERT} & &\\
		SyntaxSQLNet(BERT) & 25.0\% & 35.8\%(\textbf{+10.8}) \\ \hline
	\end{tabular}
	\caption{\label{tab:lf_results} Exact matching accuracy on development set. The header `SQL' means that the approaches are learned to generate SQL, while the header `SemQL' indicates that they are learned to generate \lf{} queries.}
\end{table}
To investigate the effectiveness of \lf{}, we alter the baseline approaches and let them learn to generate \lf{} queries rather than SQL queries.
As shown in Table~\ref{tab:lf_results}, there are at least $6.6\%$ and up to $14.4\%$ absolute improvements on accuracy of exact matching on the development set.
For example, when SyntaxSQLNet is learned to generate \lf{} queries instead of SQL queries, it registers $8.6\%$ absolute improvement and even outperforms SyntaxSQLNet(augment) which performs large-scale data augmentation.
The relatively limited improvement on TypeSQL and SQLNet is because their slot-filling based models only support a subset of \lf{} queries.
The notable improvement, on the one hand, demonstrates the effectiveness of \lf{}.
On the other hand, it shows that designing an intermediate representations to bridge NL and SQL is promising in Text-to-SQL.

\begin{table}[t!]
	\small
	\centering
	\begin{tabular}{lcc}
		\hline 
		\multicolumn{1}{l}{\textbf{Technique}} & 
		\multicolumn{1}{c}{\textbf{\approach{}}} &
		\multicolumn{1}{c}{\textbf{\approach{}(BERT)}} 
		 \\ \hline \hline
		Base model & 40.5\% & 53.9\% \\
		\hspace{2mm}+SL & 48.5\% & 60.3\% \\		
		\hspace{2mm}+SL + MEM & 51.3\% & 60.6\% \\
		\hspace{2mm}+SL + MEM + CF & 53.2\% & 61.9\% \\
		\hline
	\end{tabular}
	\caption{\label{tab:ablation_results} Ablation study results. Base model means that we does not perform schema linking (SL), memory augmented pointer network (MEM) and the coarse-to-fine framework (CF) on it.}
\end{table}
\subsection{Ablation Study}
We conduct ablation studies on \approach{} and \approach{}(BERT) to analyze the contribution of each design choice.
Specifically, we first evaluate a base model that does not apply schema linking (SL) and the coarse-to-fine framework (CF), and replace the memory augment pointer network (MEM) with the vanilla pointer network~\citep{Vinvals2015Pointer}.
Then, we gradually apply each component on the base model.
The ablation study is conducted on the development set.

Table~\ref{tab:ablation_results} presents the ablation study results. 
It is clear that our base model significantly outperforms SyntaxSQLNet, SyntaxSQLNet(
augment) and SyntaxSQLNet(BERT).
Performing schema linking (`+SL') brings about $8.5\%$ and $6.4\%$ absolute improvement on \approach{} and \approach{}(BERT).
Predicting columns in the \texttt{WHERE} clause is known to be challenging~\citep{Yavuz2018takes}.
The F1 score on the \texttt{WHERE} clause increases by $12.5\%$ when \approach{} performs schema linking.
The significant improvement demonstrates the effectiveness of schema linking in addressing the lexical problem.
Using the memory augmented pointer network (`+MEM') further improves the performance of \approach{} and \approach{}(BERT).
We observe that the vanilla pointer network is prone to selecting same columns during synthesis.
The number of examples suffering from this problem decreases by $70\%$, when using the memory augmented pointer network.
At last, adopting the coarse-to-fine framework (`+CF') can further boost performance. 

\subsection{Error Analysis}
To understand the source of errors, we analyze 483 failed examples of \approach{} on the development set.
We identify three major causes for failures:

\noindent \textbf{Column Prediction.}
We find that $32.3\%$ of failed examples are caused by incorrect column predictions based on cell values.
That is, the correct column name is not mentioned in a question, but the cell value that belongs to it is mentioned.
As the study points out~\cite{Yavuz2018takes}, the cell values of a database are crucial in order to solve this problem. 
$15.2\%$ of the failed examples fail to predict correct columns that partially appear in questions or appear in their synonym forms.
Such failures may can be further resolved by combining our string-match based method with embedding-match based methods~\citep{Krishnamurthy2017NSP} to improve the schema linking in the future. 

\noindent \textbf{Nested Query.}
$23.9\%$ of failed examples are caused by the complicated nested queries.
Most of these examples are in the Extra Hard level.
In the current training set, the number of SQL queries in Extra Hard level ($\sim$20\%) is the least, even less than the SQL queries in Easy level ($\sim$23\%).
In view of the extremely large search space of the complicated SQL queries, data augmentation techniques may be indispensable.

\noindent \textbf{Operator.}
$12.4\%$ of failed examples make mistake in the operator as it requires common knowledge to predict the correct one.
Considering the following question, `Find the name and membership level of the visitors whose membership level is higher than 4, and sort by their age from old to young', the phrase `from old to young' indicates that sorting should be conducted in descending order.
The operator defined here includes aggregate functions, operators in \texttt{WHERE} clause and the sorting orders (ASC and DESC).

Other failed examples cannot be easily categorized into one of the categories above.
A few of them are caused by the incorrect \texttt{FROM} clause, because the ground truth SQL queries join those tables without foreign key relations defined in the schema.
This violates our assumption that the definition of a database schema should be precise and complete.


When incorporated with BERT, $30.5\%$ of failed examples are fixed.
Most of them are in the category Column Prediction and Operator, but the improvement on Nested Query is quite limited.
\section{Discussion}
\label{sec:discussion}

\textbf{Performance Gap.}
There exists a performance gap on \approach{} between the development set and the test set, as shown in Table~\ref{tab:main_results}.
Considering the explosive combination of nested queries in SQL and the limited number of data (1034 in development, 2147 in test), the gap is probably caused by the different distributions of the SQL queries in Hard and Extra level.
To verify the hypothesis, we construct a pseudo test set from the official training set.
We train \approach{} on the remaining data in the training set and evaluate them on the development set and the pseudo test set, respectively.
We find that even though the pseudo set has the same number of complicated SQL queries (Hard and Extra Hard) with the development set, there still exists a performance gap.
Other approaches do not exhibit the performance gap because of their relatively poor performance on the complicated SQL queries.
For example, SyntaxSQLNet only achieves $4.6\%$ on the SQL queries in Extra Hard level on test set.
Supplementary material provides detailed experimental settings and results on the pseudo test set.

\noindent \textbf{Limitations of \lf{}.}
There are a few limitations of our intermediate representation.
Firstly, it cannot support the self join in the \texttt{FROM} clause of SQL.
In order to support the self join, the variable mechanism in lambda calculus~\citep{carpenter1997type} or the scope mechanism in Discourse Representation Structure~\citep{Hans1993DST} may be necessary.
Secondly, \lf{} has not completely eliminated the mismatch between NL and SQL yet.
For example, the \texttt{INTERSECT} clause in SQL is often used to express disjoint conditions.
However, when specifying requirements, end users rarely concern about whether two conditions are disjointed or not.
Despite the limitations of \lf{}, experimental results demonstrate its effectiveness in Text-to-SQL.
To this end, we argue that designing an effective intermediate representation to bridge NL and SQL is a promising direction to being there for complex and cross-domain Text-to-SQL.
We leave a better intermediate representation as one of our future works.
\section{Related Work}
\label{sec:related}
\noindent \textbf{Natural Language Interface to Database.}
The task of Natural Language Interface to Database (NLIDB) has received significant attention since the 1970s~\citep{Warren1981EasilyAS, androutsopoulos1995natural, Popescu2004MNL, Hallett2006GQR, Giordani2012GSQ}.
Most of the early proposed systems are hand-crafted to a specific database~\citep{Warren1982EEA, Woods1986Lunar, Hendrix1978DNL}, making it challenging to accommodate cross-domain settings.
Later work focus on building a system that can be reused for multiple databases with minimal human efforts~\citep{Grosz1987TED, Androutsopoulos1993MASQUE,  Tang2000ACD}.
Recently, with the development of advanced neural approaches on Semantic Parsing and the release of large-scale, cross-domain Text-to-SQL benchmarks such as WikiSQL~\citep{ZhongSeq2SQL2017} and Spider~\citep{Spider2018}, there is a renewed interest in the task~\citep{Sqlnet2017, Iyer2017LNS, bo2018Table, DialSQL2018, Typesql2018, Syntaxsqlnet2018, wang2018robust, Finegan2018text2sql, Noval2019}.
Unlike these neural approaches that end-to-end synthesize a SQL query, \approach{} first synthesizes a \lf{} query and then infers a SQL query from it.

\noindent \textbf{Intermediate Representations in NLIDB.}
Early proposed systems like as LUNAR~\citep{Woods1986Lunar} and MASQUE~\citep{Androutsopoulos1993MASQUE} also propose intermediate representations (IR) to represent the meaning of questions and then translate it into SQL queries.
The predicates in these IRs are designed for a specific database, which sets \lf{} apart.
\lf{} targets a wide adoption and no human effort is needed when it is used in a new domain.
\citet{Li2014CIN} propose a query tree in their NLIDB system to represent the meaning of a question and it mainly serves as an interaction medium between users and their system.

\noindent \textbf{Entity Linking.}
The insight behind performing schema linking is partly inspired by the success of incorporating entity linking in knowledge base question answering and semantic parsing~\citep{Yih2016Value, Krishnamurthy2017NSP, Typesql2018, Herzig2018Decouple, Kolitsas2018NEL}.
In the context of semantic parsing, \citet{Krishnamurthy2017NSP} propose a neural entity linking module for answering compositional questions on semi-structured tables. 
TypeSQL~\citep{Typesql2018} proposes to utilize type information to better understand rare entities and numbers in questions.
Similar to TypeSQL, \approach{} also recognizes the columns and tables mentioned in a question. 
What sets \approach{} apart is that \approach{} assigns different types to the columns based on how they are mentioned in the question.
\section{Conclusion}
\label{sec:conclusion}

We present a neural approach \lf{}  for complex and cross-domain Text-to-SQL, aiming to address the lexical problem and the mismatch problem with schema linking and an intermediate representation.
Experimental results on the challenging Spider benchmark demonstrate the effectiveness of \approach{}.
\section*{Acknowledgments}
We would like to thank Bo Pang and Tao Yu for evaluating our submitted models on the test set of the Spider benchmark.
Ting Liu is the corresponding author.

\bibliography{acl2019}
\bibliographystyle{acl_natbib}

\section{Supplemental Material}
\label{sec:supplemental}

\subsection{Examples of \lf{} Query}
\begin{figure*}[t!]
	\centering
	\includegraphics[scale=0.36]{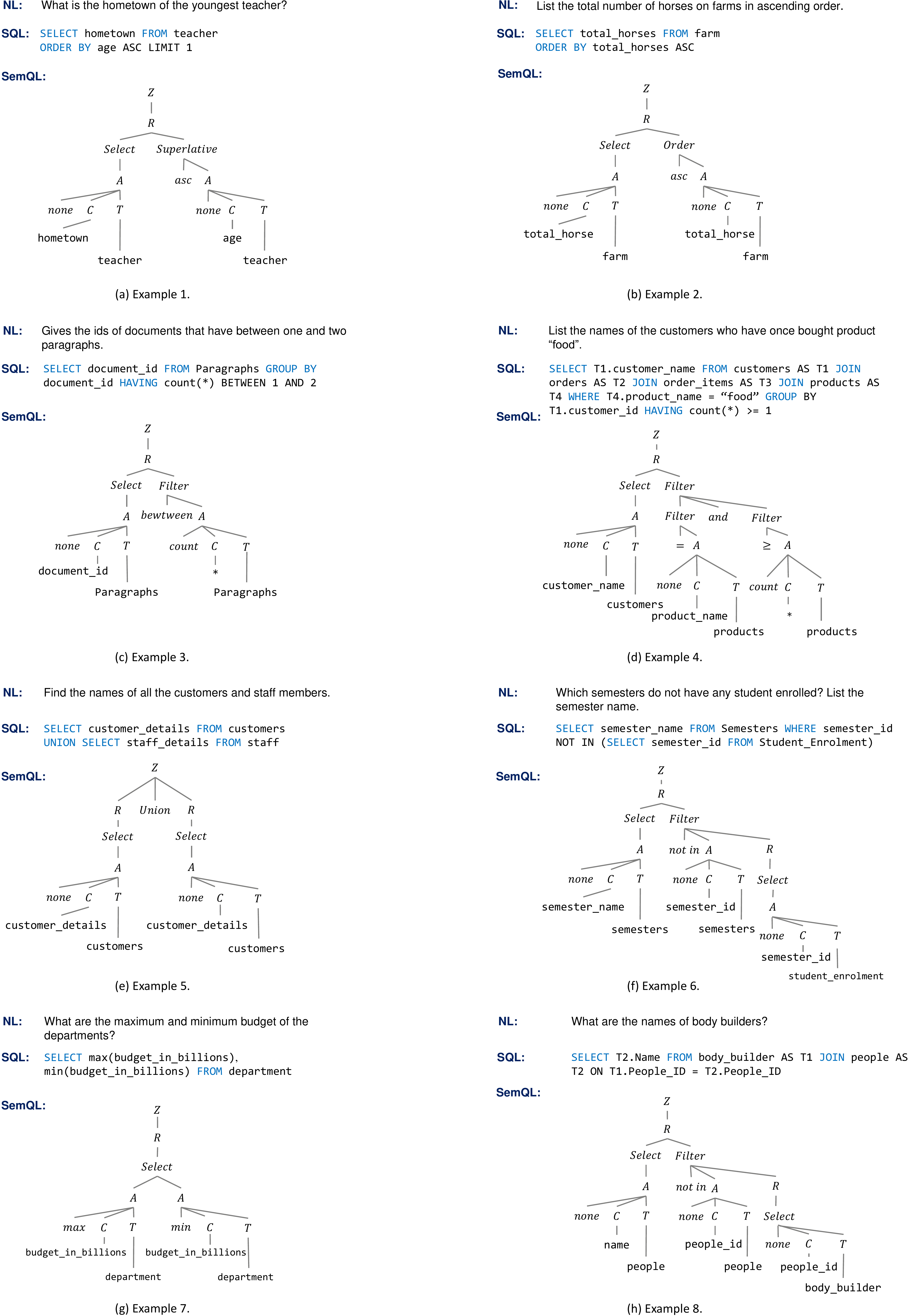}
	\caption{
		\label{fig:logical_forms_example} Examples of SemQL Query.}
\end{figure*}

Figure \ref{fig:logical_forms_example} presents more examples of \lf{} queries.

\subsection{Inference of SQL Query}
To infer a SQL query from a \lf{} query, we traverse the tree-structured \lf{} query in pre-order and map each tree node to the corresponding SQL query components according to the production rule applied to it. 

The production rule applied to the \textit{Z} node denotes whether the SQL query has one of the following components, \texttt{UNION}, \texttt{EXCEPT} and \texttt{INTERSECT}. 
The \textit{R} node stands for the start of a single SQL query.
The production rule applied to $R$ denotes whether the SQL query has a \texttt{WHERE} clause and \texttt{ORDERBY} clause.
The production rule applied to a \textit{Select} node denotes how many columns does the \texttt{SELECT} clause has. 
Each \textit{A} node denotes a column/aggregate function pair.
Specifically, nodes under \textit{A} denote the aggregate function, the column name and the table name of the column.
The subtrees under nodes \textit{Superlative} and \textit{Order} are mapped to the \texttt{ORDERBY} clause in the SQL query.
The production rules applied to \textit{Filter} denote different condition operators in SQL query, e.g. \texttt{and}, \texttt{or}, \texttt{>}, \texttt{<}, \texttt{=}, \texttt{in}, \texttt{not in} and so on.
If there is a \textit{A} node under the \textit{Filter} node and its aggregate function is not $None$, it will be filled in the \texttt{HAVING} clause, otherwise in the \texttt{WHERE} clause.
If there is a \textit{R} node under the \textit{Filter} node, we will repeat the process recursively on the \textit{R} node and return a nested SQL query.
The \texttt{FROM} clause is generated from the selected tables in the \lf{} query by identifying the shortest path that connects these tables in the schema (Database schema can be formulated as an undirected graph, where vertex are tables and edges are relations among tables).
At last, if there exists an aggregate function applied on a column in the \lf{} query, there should be \texttt{GROUPBY} clause in the SQL query.
The column to be grouped by occurs in the \texttt{SELECT} clause in most cases, or it is the primary key of a table where an aggregate function is applied on one of its columns.

\subsection{Transforming SQL to SemQL}

To generate a \lf{} query from a SQL query, we first initialize a \textit{Z} node.
If the SQL query has one of the components \texttt{UNION}, \texttt{EXCEPT} and \texttt{INTERSECT}, we attach the corresponding keywords and two \textit{R} nodes under \textit{Z}, otherwise a single \textit{R} node.
Then, we attach a \textit{Select} node under \textit{R}, and the number of columns in \texttt{SELECT} clause determines the number of \textit{A} nodes under the \textit{Select} node.
If an \texttt{ORDERBY} clause in a SQL query contains a \texttt{LIMIT} keyword, it will be transformed into a \textit{Superlative} node, otherwise a \textit{Order} node.
Next, the sub-tree of \textit{Filter} node is determined by the condition in \texttt{WHERE} and \texttt{HAVING} clause. 
If it has a nested query in \texttt{WHERE} clause or \texttt{HAVING} clause, we process the subquery recursively.
For each column in a SQL query, we attach its aggregate function node, a \textit{C} node and a \textit{T} node under \textit{A}. 
Node \textit{C} attaches the column name and node \textit{T} attaches its table name.
For the special column `\texttt{*}', if there is only one table in the \texttt{FROM} clause that does not belongs to any column, we assign it the column `\texttt{*}', otherwise, we label the table name of `\texttt{*}' manually.
If a table in \texttt{FROM} clause is not assigned to any column, it will be transformed into a subtree under a $Filter$ node with $in$ condition.
In this way, a SQL query can be successfully transformed into a \lf{} query.

\subsection{Coarse-to-Fine Framework}

\begin{figure}[t]
	\centering
	\includegraphics[scale=0.50]{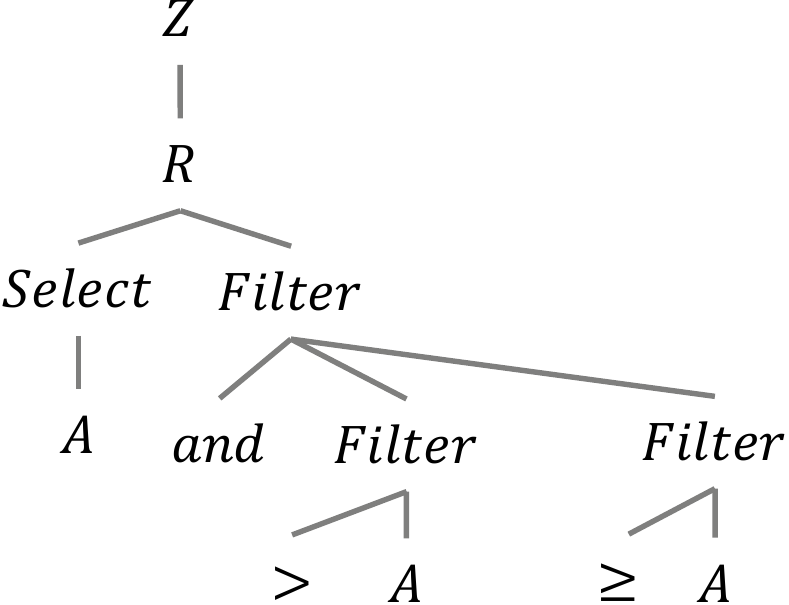}
	\caption{
		\label{fig:lf_skeleton} The skeleton of the \lf{} query presented in Figure~\ref{fig:lf_example}}
\end{figure}

\begin{figure*}[h!]
	\centering
	\includegraphics[scale=0.33]{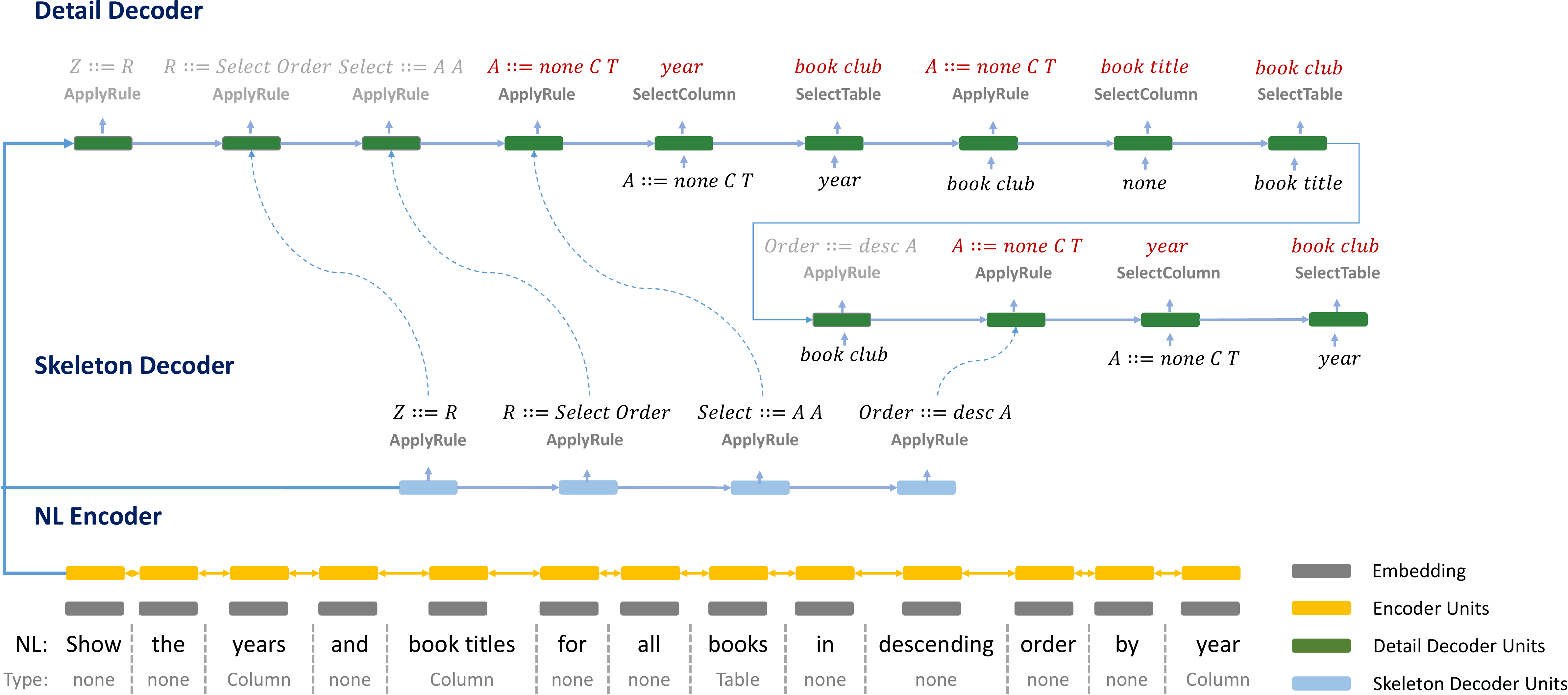}
	\caption{
		\label{fig:coarse_2_fine} An overview of the coarse-to-fine framework to synthesize \lf{}.}
\end{figure*}

The skeleton of a \lf{} query is obtained by removing all nodes under each \textit{A} node.
Figure~\ref{fig:lf_skeleton} shows the skeleton of the \lf{} query presented in Figure~\ref{fig:lf_example}.

Figure~\ref{fig:coarse_2_fine} depicts the coarse-to-fine framework to synthesize a \lf{} query. 
In the first stage, a skeleton decoder outputs the skeleton of a \lf{} query.
Then, a detail decoder fills in the missing details in the skeleton by selecting columns and tables.
The probability of generating a \lf{} query $y$ in the coarse-to-fine framework is formalized as follows.
\begin{equation*}
\small
\begin{aligned}
p(y|x, s)&\!=\! p(q|x, s)p(y|x, s, q) \\
p(q|x, s) &\!=\! \prod^{T_s}_{i=1}p(a_i\!=\!  \textsc{ApplyRule}[r]|x, s, a_{<i}) \\
p(y|x, s, q) &\!=\! \prod^{T_c}_{i=1} [\lambda_i p(a_i\!=\! \textsc{SelectColumn}[c]|x, s, q, a_{<i})\\ & +  (1 \!-\! \lambda_i) p(a_i\!=\! \textsc{SelectTable}[t]|x, s, q, a_{<i})] \\
\end{aligned}	
\end{equation*}
where $q$ denotes the skeleton. $\lambda_i = 1$ when the $i$th action type is SelectColumn, otherwise $\lambda_i = 0$.

At training time, our model is optimized by maximizing the log-likelihood of the ground true action sequences:
\begin{equation*}
\begin{aligned}
max\sum_{(x, s, q, y) \in \mathcal{D}} \log p(y|x, s, q) + \gamma\log p(q|x, s)
\end{aligned}
\end{equation*}
where $\mathcal{D}$ denotes the training data and $\gamma$ represents the scale between $\log p(y|x, s, q)$ and $ \log p(q|x, s)$. $\gamma$ is set to 1 in our experiment.

\subsection{BERT}
\begin{figure*}[h!]
	\centering
	\includegraphics[scale=0.30]{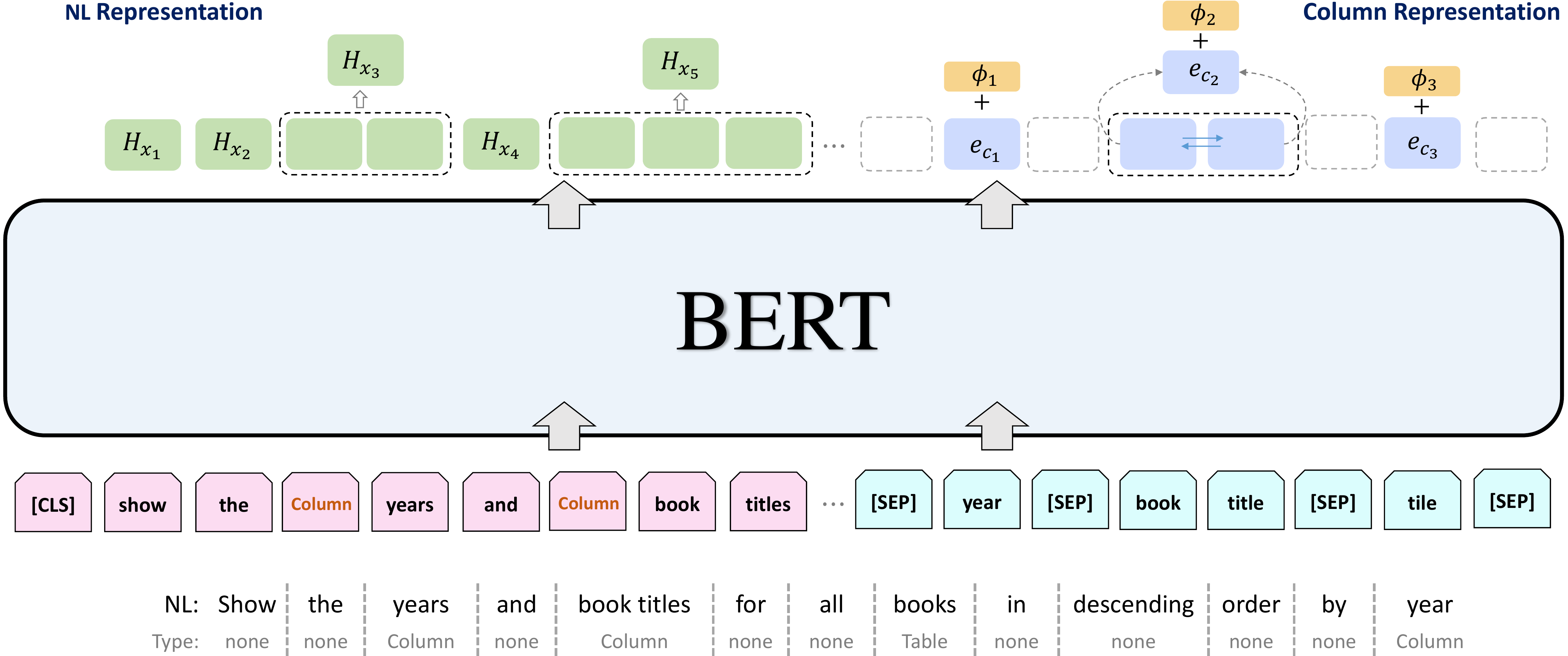}
	\caption{
		\label{fig:bert} Encoding a question and column names with BERT.}
\end{figure*}
Figure~\ref{fig:bert} depicts the architecture of the BERT encoder.

\subsection{Analysis on the Performance Gap between the Development set and the Test set}
\begin{table}[h]
	\small
	\centering
	\begin{tabular}{lcccc}
		\hline 
		\small{\textbf{\multirow{2}*{Dataset}}} & 
		\small{\textbf{\multirow{2}*{Easy}}}   & 
		\small{\textbf{\multirow{2}*{Medium}}}  & 
		\small{\textbf{\multirow{2}*{Hard}}}  & 
		\small{\textbf{Extra}} \\
		&&&&\small{\textbf{Hard}}  \\
        \hline \hline
		\small{Pseudo Test A}  & 24.2\% & 44.5\% & 14.4\% & 16.9\%\\ 
		\small{Pseudo Test B}  & 22.7\% & 44.1\% & 16.7\% & 16.5\% \\ 
		\small{Pseudo Test C}  & 24.7\% & 37.1\% & 22.9\% & 15.3\% \\ 
		\textbf{\small{Development}}  & \textbf{24.1\%} & \textbf{42.5\%} & \textbf{16.8\%} & \textbf{16.4\%}  \\
		 \hline
	\end{tabular}
	\caption{\label{tab:hardness_split} The hardness distribution of the pseudo test A, the pseudo test B, the pseudo test C and the development set.}
\end{table}

\begin{table*}[h!]
\small
\centering
\begin{tabular}{l|ccc|ccc}
\hline
\multicolumn{1}{l|}{\multirow{2}{*}{\textbf{Approach}}} & \multicolumn{3}{c|}{\textbf{Dev}}                                                                                 & \multicolumn{3}{c}{\textbf{Pseudo Test A}}                                                                                \\ \cline{2-7} 
\multicolumn{1}{l|}{}                                   & \multicolumn{1}{c}{\textbf{All}} & \multicolumn{1}{c}{\textbf{Hard}} & \multicolumn{1}{c|}{\textbf{Extra Hard}} & \multicolumn{1}{c}{\textbf{All}} & \multicolumn{1}{c}{\textbf{Hard}} & \multicolumn{1}{c}{\textbf{Extra Hard}} \\ \hline \hline
SyntaxSQLNet    & 17.4\%    & 15.5\%    & 2.9\%  & 16.9\%       & 11.0\%    & 2.6\%      \\
\hspace{2mm} +BERT +\lf{}  & 34.5\%  & 30.5\%    & 17.6\%        & 30.2\%    & 24.5\%   & 15.7\% \\ \hline
\end{tabular}
	\caption{\label{tab:syntax_base} Exact matching accuracy on the development set and the pseudo test A set.}
\end{table*}

To test our hypothesis that the performance gap is caused by the different distribution of the SQL queries in Hard and Extra Hard level, we first construct a pseudo test set from the official training set of Spider benchmark.
Then, we conduct further experiment on the pseudo test set and the official development set.
Specifically, we sample 20 databases from the training set to construct a pseudo test set, which has the same hardness distributions with the development set.
Then, we train \approach{} on the remaining training set, and evaluate it on the development set and the pseudo test set, respectively.
We sample the pseudo test set from the training set for three times and obtain three pseudo test sets, namely, pseudo test A, pseudo test B and pseudo test C.
They contain 1134, 1000 and 955 test data respectively.

\begin{figure*}[h!]
    \centering
    \begin{subfigure}{.3\linewidth}
		\centering
		\includegraphics[scale=0.33]{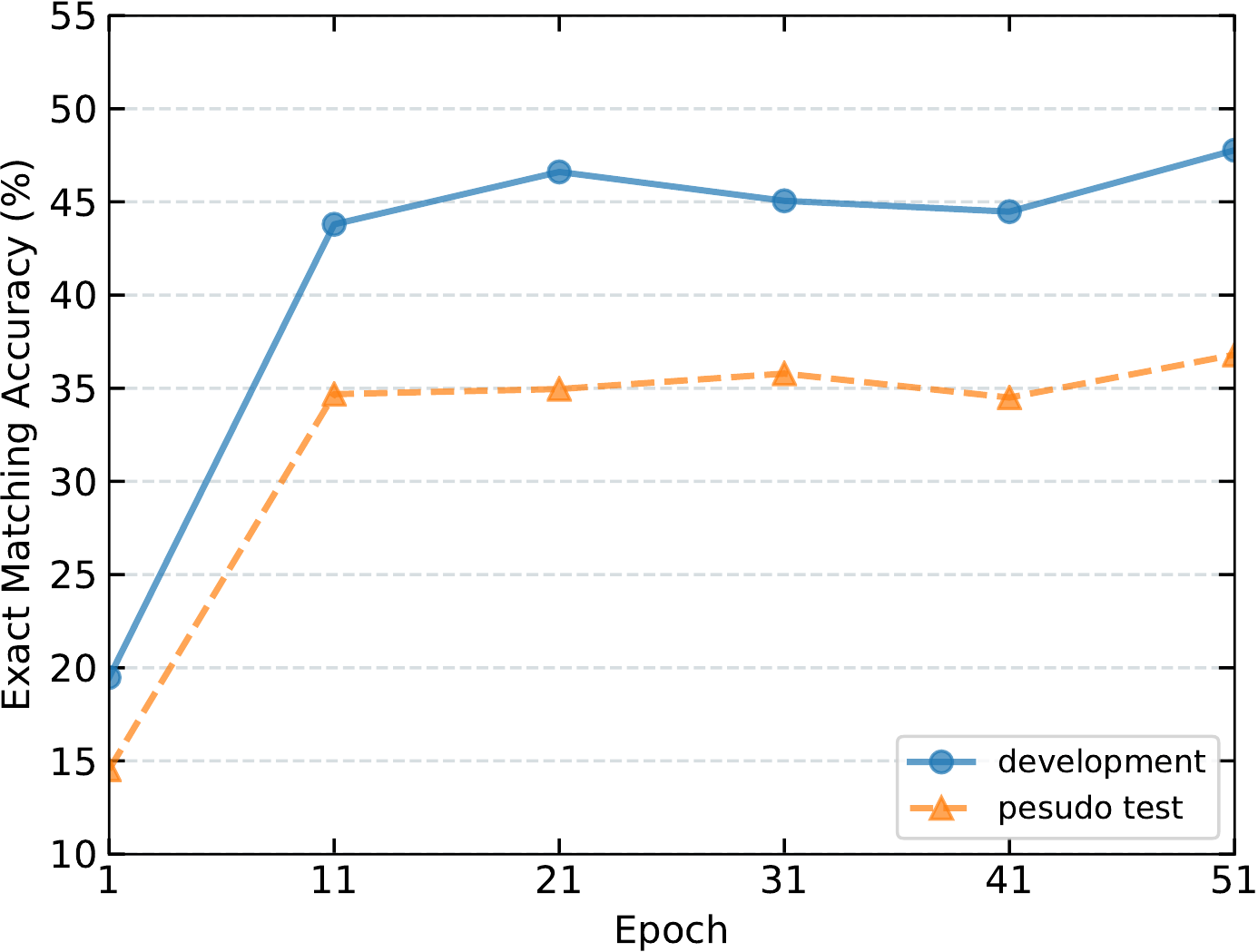}
	   \caption{\label{subfig:pseudoa}Pseudo Test A}
	\end{subfigure}\hfill
	\begin{subfigure}{.3\linewidth}
		\centering
		\includegraphics[scale=0.33]{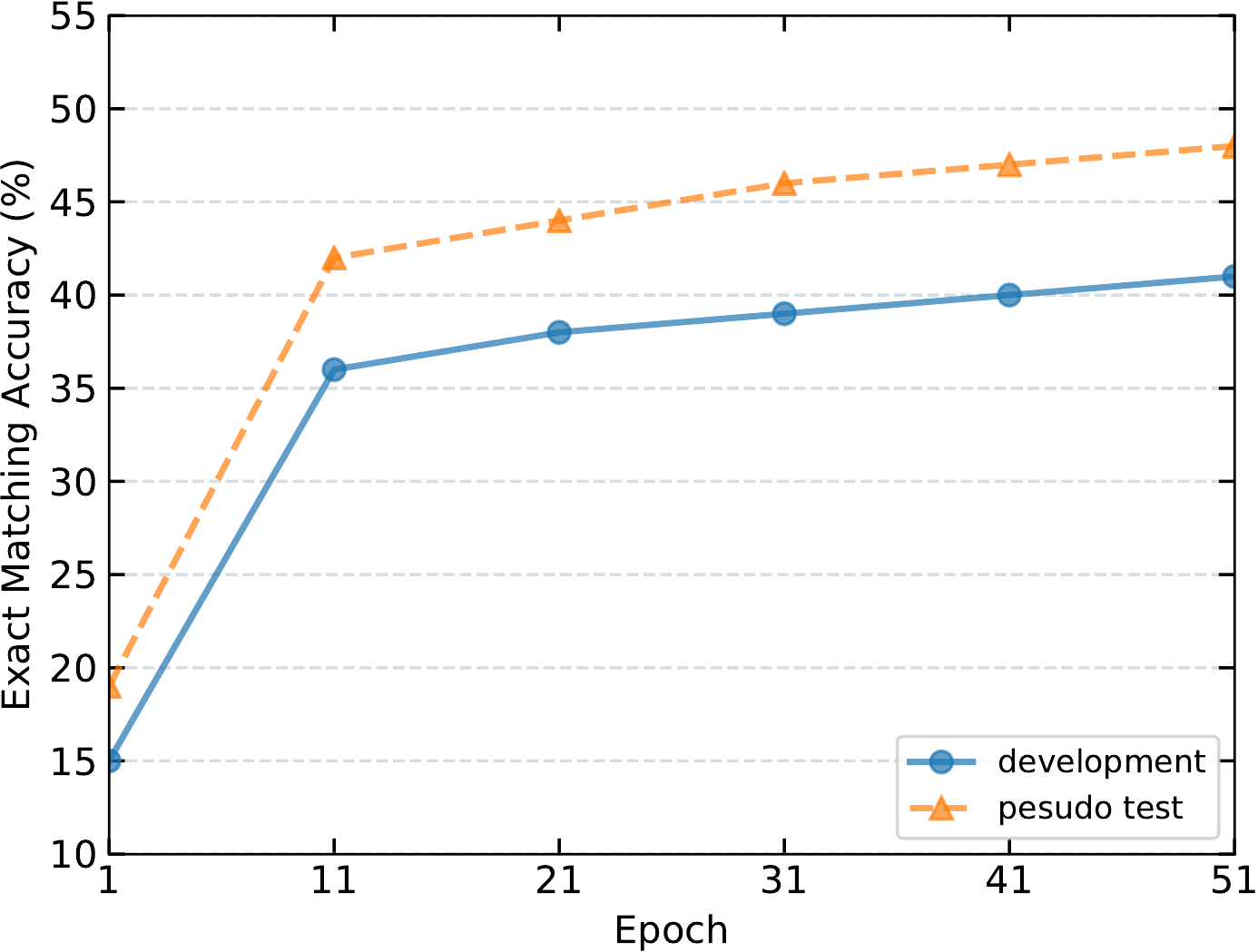}
		\caption{\label{subfig:pseudob}Pseudo Test B}
	\end{subfigure}\hfill
	\begin{subfigure}{.3\linewidth}
		\centering
		\includegraphics[scale=0.33]{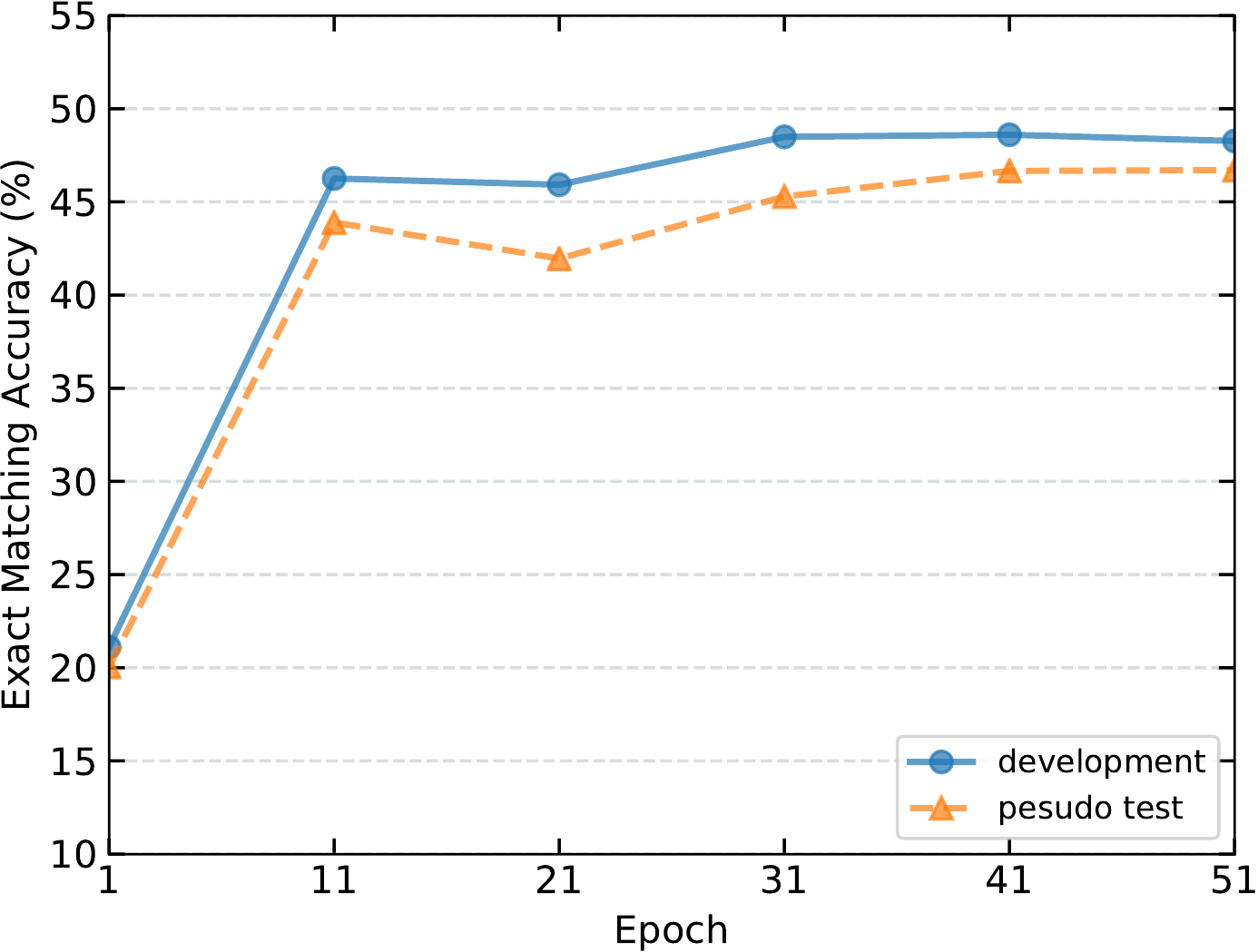}
		\caption{\label{subfig:pseudoc}Pseudo Test C}
	\end{subfigure}\hfill
    \caption{Exact matching accuracy of \approach{} on development set and pseudo test sets.}
    \label{fig:dev_test_performance}
\end{figure*}

Table~\ref{tab:hardness_split} presents the hardness distribution of the three pseudo test sets and the official development set.
Figure~\ref{fig:dev_test_performance} presents the exact matching accuracy of SQL on the development set and three pseudo tests set after each epoch during training.
\approach{} performs competitively on the development set and the pseudo set C (Figure~\ref{subfig:pseudoc}), but there exists a clear performance gap on the pseudo test A and B (Figure~\ref{subfig:pseudoa} and Figure~\ref{subfig:pseudob}).
Although the hardness distributions among the development set and the three pseudo sets are nearly the same, the data distribution still has some difference, which results in the performance gap.

We further study the performance gap of SyntaxSQLNet on the development set and the pseudo test A.
As shown in Table~\ref{tab:syntax_base},
SyntaxSQLNet achieves $16.9\%$ on the development set and $17.4\%$ on the pseudo test A.
When incorporating BERT and learning to synthesizing \lf{}, SyntaxSQLNet(BERT,\lf{})
achieves $34.5\%$ on the development set and $30.2\%$ on the pseudo test A, exhibiting a clear performance gap ($4.3\%$).
SyntaxSQLNet(BERT, \lf{}) significantly outperforms SyntaxSQLNet in the Hard and Extra Hard level.
The experimental results show that when SyntaxSQLNet performs better in the Hard and Extra Hard level, the performance gap will be larger, since that the performance gap is caused by the different data distributions.

\end{document}